\documentclass[10pt,letterpaper]{article}

\usepackage{arxiv/preprint}

\usepackage{amsmath}
\usepackage{amssymb}
\usepackage{booktabs}
\usepackage{multirow}
\usepackage{xspace}
\usepackage{algorithm}
\usepackage{algpseudocode}
\usepackage{wrapfig}

\graphicspath{{./}}

\definecolor{oursgreen}{RGB}{223,242,223}
\newcommand{\pmval}[2]{#1\,\textmd{\color{black!55}\scriptsize$\pm$\,#2}}

\newcommand{\modelwordmark}{\texorpdfstring{\mbox{%
  \textcolor{ppBlue}{H}%
  \textcolor{ppBlue!71!ppAccent}{I}%
  \textcolor{ppBlue!43!ppAccent}{G}%
  \textcolor{ppBlue!14!ppAccent}{e}%
  \textcolor{ppAccent!86!ppPink}{n}%
  \textcolor{ppAccent!57!ppPink}{N}%
  \textcolor{ppAccent!29!ppPink}{T}%
  \textcolor{ppPink}{O}}}{HIGenNTO}}

\makeatletter

\newcommand{\model}{\text{HIGenNTO}}

\newcommand*{\etc}{%
    \@ifnextchar{.}%
        {\textit{etc}}%
        {\textit{etc.}\@\xspace}%
}
\makeatother

\newcommand{\taskone}{\texttt{Climbing Stairs}}
\newcommand{\tasktwo}{\texttt{Descending Stairs}}
\newcommand{\taskthree}{\texttt{Sit on a Chair}}
\newcommand{\taskfour}{\texttt{Standing Up}}
\newcommand{\taskfive}{\texttt{Pick and Place Box}}
\newcommand{\tasksix}{\texttt{Hug}}
\newcommand{\taskseven}{\texttt{HighFive}}
\newcommand{\taskeight}{\texttt{Pick and Place on Table}}
\newcommand{\tasknine}{\texttt{Slalom Walking}}
\newcommand{\taskten}{\texttt{Step Up/Down}}
\newcommand{\taskeleven}{\texttt{Crawl Under}}
\newcommand{\tasktwelve}{\texttt{Duck Under}}
\newcommand{\taskthirteen}{\texttt{Squeeze Through Gap}}
\newcommand{\taskfourteen}{\texttt{Push a Box}}

\hypersetup{
  pdftitle={HIGenNTO: Scalable Humanoid Interaction Generation via Noise-Space Trajectory Optimization},
  pdfauthor={Lalit Jayanti, Kashu Yamazaki, Yuto Shibata, Kotaro Amaya, Katerina Fragkiadaki},
}

\begin{document}

\begin{headercard}
\cardtitle{\modelwordmark: Scalable Humanoid Interaction Generation\\ via Noise-Space Trajectory Optimization}

\cardauthors{Lalit Jayanti\textsuperscript{1}, \ Kashu Yamazaki\textsuperscript{1}, \ Yuto Shibata\textsuperscript{2,3}, \ Kotaro Amaya\textsuperscript{2,3}, \ Katerina Fragkiadaki\textsuperscript{1}}
\cardaffil{\textsuperscript{1}Carnegie Mellon University \quad
  \textsuperscript{2}Keio AI Research Center \quad
  \textsuperscript{3}Keio University}
\cardrule
\renewcommand{\model}{\modelwordmark}%
\cardabstract{Humanoid robots can acquire complex skills by imitating kinematic humanoid motion references, yet reliable references for contact-rich interactions remain difficult to obtain: motion capture deteriorates under occlusion and close physical contact, while retargeting introduces additional contact and geometric inconsistencies. We present \model{}, a framework that synthesizes humanoid--scene interaction motion references by optimizing the initial noise of a pretrained text-conditioned motion model under sparse spatiotemporal and scene constraints. The same formulation satisfies desired contacts, avoids collisions, and maintains stable support while retaining the prior's realism and temporal coherence, generating interaction motions from scratch and composing long-horizon behaviors stage-wise. Across robot--environment and robot--object tasks, \model{} produces motions that can be executed by tracking policies in simulation and used to train depth-conditioned visuomotor policies operating solely from onboard sensing. We deploy these policies on a Unitree G1 across four contact-rich tasks. Finally, the task specifications themselves can be written by a coding agent, which proposes interaction tasks and compiles them into prompt, constraint, and scene programs, authoring three of our eight evaluated tasks and four further behaviors. Together, these results establish a scalable path from high-level task descriptions to physically executable humanoid interactions.
}
\vspace{7pt}
\cardmeta{Website}{\url{https://higennto.github.io}}
\end{headercard}

\section{Introduction}
\label{sec:intro}
\begin{figure}[t]
    \centering
    \includegraphics[width=\linewidth]{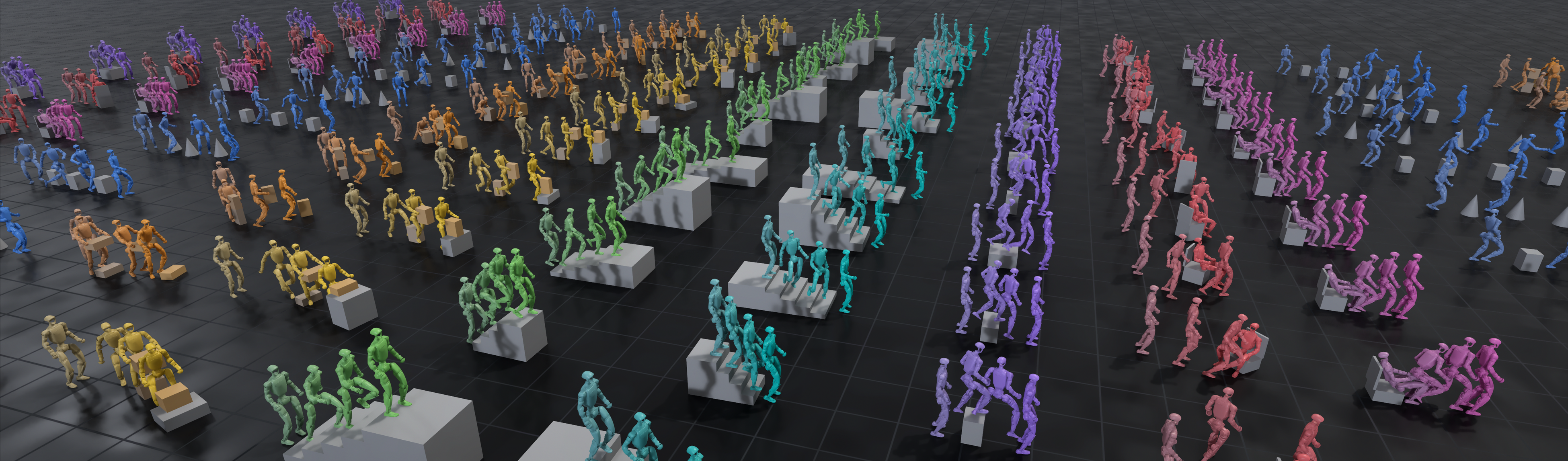}
    \caption{\textbf{\model{} synthesizes humanoid--scene interaction motion references through constrained optimization over a generative motion prior.} Stair traversal, sitting and standing, and box pickup and placement across randomized scenes, without any motion capture of joint human--scene interaction.}
    \label{fig:teaser}
\end{figure}

Humanoid robots have acquired remarkably complex skills by imitating kinematic 3D trajectories recovered from motion capture or monocular video~\cite{peng2018deepmimic,luo2026sonic,chen2025gmt}, and inherit the limitations of that reference data. Most large-scale datasets record the body in isolation, omitting the scene geometry, object motion, and contacts that constitute interaction~\cite{mahmood2019amass,guo2022humanml3d}. Interaction is also the hardest case to capture: occlusion and close contact degrade pose estimation where precision matters most, so even datasets built for it~\cite{interx,omomo} contain missed contacts, interpenetrations, and temporal inconsistencies, which retargeting compounds. 

We introduce \model{} (Humanoid Interaction Generation via Noise-Space Trajectory Optimization), which synthesizes contact-rich references without interaction motion capture (Figure~\ref{fig:teaser}). \textbf{We cast interaction synthesis as constrained optimization over a generative motion prior}: \model{} optimizes the initial noise of a pretrained text-conditioned motion model subject to sparse spatiotemporal constraints and differentiable scene objectives that specify what text and keyframes cannot: which body part contacts which object, where and when, and which regions to keep clear. Gradients traverse the whole sampling chain, so the trajectory adapts to contact, collision, support, and task geometry while staying on the prior's learned manifold. A single formulation yields motions from a task specification and scene geometry alone, and can extend past the pretrained motion prior's prediction horizon.

Optimizing diffusion noise has previously enabled motion editing and control of human characters, including navigation in a static scene~\cite{karunratanakul2024dno,dartcontrol}. We extend this mechanism to humanoid interaction synthesis, where contacts must not only be avoided but also deliberately established and maintained with support surfaces, manipulated objects, and other agents.

The specification for generating motions itself can be automated: from a short brief and the corpus annotations, a coding agent sizes the scene and writes the program to generate the spatiotemporal constraints. Further, these generated references can be executed by tracking policies and used to train depth-conditioned
visuomotor policies operating solely from onboard sensing. 

\noindent \textbf{Contributions.}
\textbf{(1) Interaction synthesis as optimization in noise space.} We make a pretrained, scene-unaware motion prior contact- and geometry-aware without retraining it, achieving lower scene penetration across all three interaction categories and lower object-task hand-target error than both kinematic-constraint conditioning and classifier guidance.
\textbf{(2) References that physical policies execute.} Across eight tasks, scene-aware teachers track the generated motion and distill into depth-only students that complete $57.4$--$98.3\%$ of motions, where methods based on scene-unaware tracking formulations~\cite{luo2026sonic} largely fail.
\textbf{(3) Task specification by a coding agent.} Agent-written programs supply three of our eight evaluated tasks and four further behaviors, with no per-task demonstration and no hand-tuned constraints.

\section{Related Work}
\paragraph{Human Motion Data and Human-to-Humanoid Retargeting.}
Large-scale motion datasets~\cite{mahmood2019amass,guo2022humanml3d,lafan} underpin humanoid motion imitation and control~\cite{peng2018deepmimic,peng2018sfv,cheng2024expressive}, with collection scaled by video reconstruction~\cite{zhou2024motionx,allshire2025videomimic}. Most model humans in isolation, and those that do model interactions with the scene~\cite{wang2025skillmimic,wang2023physhoi,hassan2021sceneaware,zhang2022couch,interx,humanx,mao2024humanoidx} still yield penetrations, contact inconsistencies, and implausible artifacts. Retargeting these motions to robots is an effective route to agile behaviors~\cite{peng2018deepmimic,araujo2025gmr,serifi2024robot}, via inverse kinematics, trajectory optimization, and learned retargeting~\cite{choi2019naturalmotion,gomes2019movement,choi2021selfsupervised,araujo2025gmr}, including interaction-aware objectives~\cite{yang2025omniretarget,weng2025hdmi,xu2025intermimic,tessler2025maskedmanipulator,cheynel2025reconform,jang2024geometryaware}, but all depend on high-quality interaction motion that remains hard to obtain at scale. We instead synthesize such references from generative motion priors under sparse, semantically grounded constraints.

\paragraph{Generative Motion Priors for Humanoid Control.}
Diffusion-based motion models generate realistic, controllable motion conditioned on text, goals, or sparse constraints~\cite{tevet2023human,kimodo,shafir2024priormdm}, and several works pair them with RL tracking controllers for physical realism at execution~\cite{tevet2024closd,serifi2024robot,xu2025parc}. Adversarial motion priors such as AMP~\cite{peng2021amp} and ASE~\cite{peng2022ase} instead regularize policies toward realistic motion distributions without explicit trajectory tracking.

\paragraph{Controllable and Constraint-Guided Motion Generation.}
Diffusion models achieve spatial and semantic control through training-time conditioning on paired control signals such as trajectories or sparse constraints~\cite{tevet2023human,shafir2024priormdm,xie2024omnicontrolcontrol,karunratanakul2023guidedmotion,kimodo}, through inference-time guidance by gradients of rewards, constraints, or value functions during denoising~\cite{janner2022planning,wang2023diffusionpoliciesexpressivepolicy,zheng2025diffusionbasedplanningautonomousdriving}, or by optimizing the initial noise itself, backpropagating a differentiable objective through the entire sampling chain. We adopt the last: DNO~\cite{karunratanakul2024dno} establishes it for editing, refinement, and obstacle avoidance, and DartControl~\cite{dartcontrol} applies it to human--scene contact. Both operate on kinematic human characters against static geometry; we target humanoid robots, add objectives for contact held with movable objects and other agents, and train physical policies on the resulting references.

\section{Method}
\label{sec:method}

\begin{figure}[t]
    \centering
    \includegraphics[width=\linewidth]{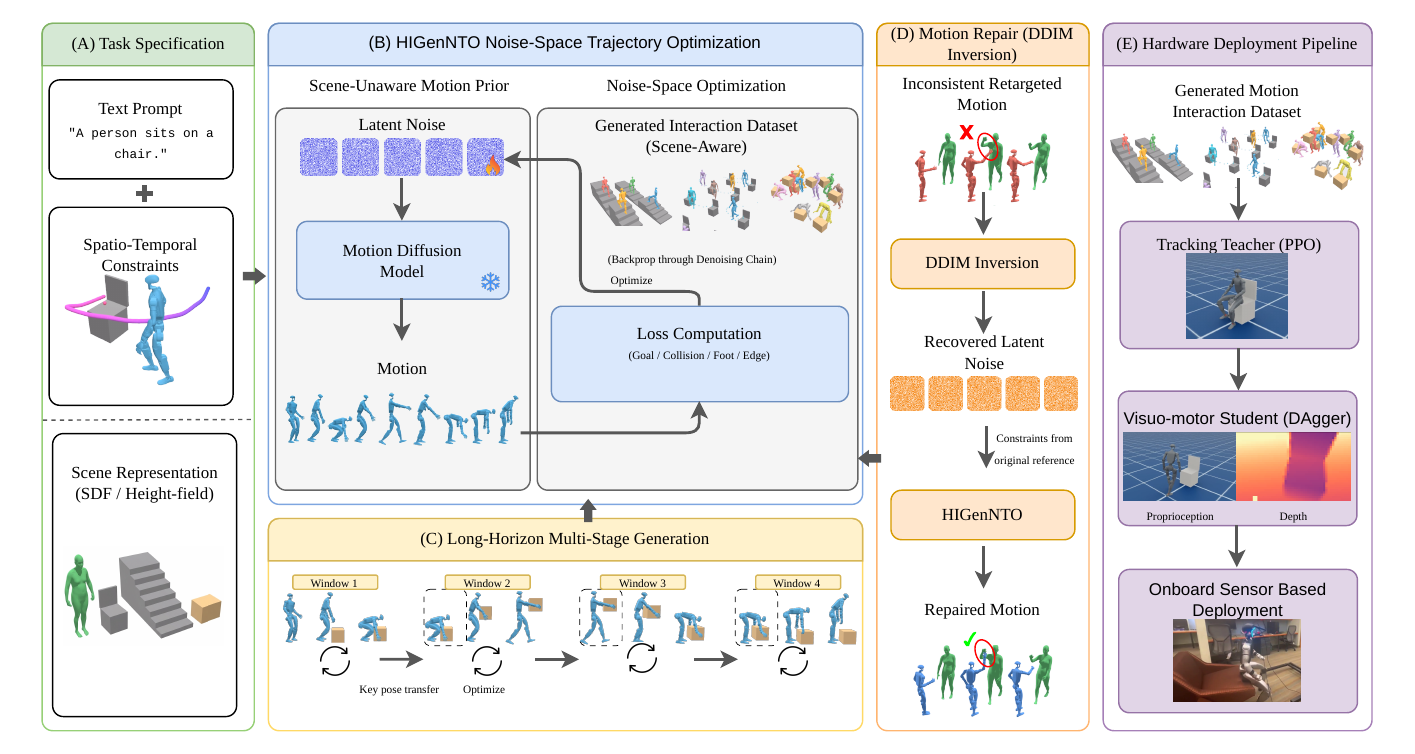}
    \caption{\textbf{System overview.} (A) \textbf{Task specification:} a text prompt, sparse spatiotemporal constraints on end-effector and root goals, and a scene representation (SDF / height-field). (B) \textbf{Noise-space optimization:} differentiable scene and contact losses (goal, collision, foot, edge) are backpropagated through the denoising chain of a frozen motion prior into its input latent noise, yielding a scene-aware interaction dataset. (C) \textbf{Long-horizon generation:} temporal windows optimized in sequence, with key poses transferred across boundaries. (D) \textbf{Motion repair:} an inconsistent retargeted motion sequence is inverted to latent noise and re-optimized. (E) \textbf{Deployment:} the dataset trains a PPO tracking teacher, distilled into a depth- and proprioception-conditioned student.}
    \label{fig:methodology}
\end{figure}

Training a scene-aware interaction generator directly would require physically consistent interaction paired with time-varying scene geometry, which is hard to collect at scale. We instead inject scene-awareness at optimization time, optimizing the latent noise of a frozen text-conditioned motion diffusion model under differentiable contact, and scene objectives (Figure~\ref{fig:methodology}). The synthesized motions stand in for motion-capture interaction data~\cite{physhsi,humanx}, supervising a privileged tracking teacher that is distilled into a policy running from onboard sensing alone.

\subsection{Scene-Aware Interaction Synthesis via Constrained Latent Optimization}
\label{sec:method:formulation}
We consider an articulated humanoid acting in a scene. A task is specified by:
(i) a text prompt $y$;
(ii) a set of sparse spatiotemporal constraints
$\mathcal{C} = \{(j_i, t_i, p_i)\}_{i=1}^{N_c}$
that constrain keypoint $j_i$ to a target 3D position $p_i \in \mathbb{R}^3$ at time $t_i$; and
(iii) a scene representation $\mathcal{S}$ given by differentiable signed-distance functions (SDFs) for static and dynamic geometry, manipulated objects, and terrain height fields. Constraints are authored per task (by hand or, as we show in Sec.~\ref{sec:exp:agent}, by a coding agent), except for repair, where contact candidates proposed from the reference are filtered by a vision-language model (Appendix~\ref{app:repair}). The goal is a motion trajectory $\mathbf{x} \in \mathbb{R}^{T \times J \times 3}$ that is semantically consistent with $y$, satisfies $\mathcal{C}$, respects scene geometry, and remains temporally smooth.

\begin{figure}[t]
    \centering
    \includegraphics[width=\linewidth]{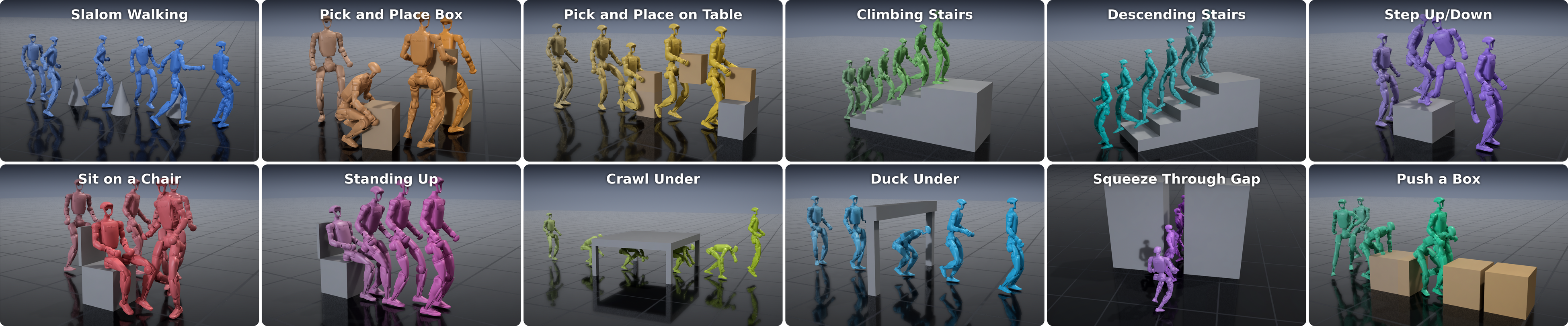}
    \caption{\textbf{Generated interaction motions across the task suite.} Each panel overlays frames from a single generated motion, labeled in-panel: the eight tasks we evaluate, together with four further behaviors whose task programs were written by the coding agent of Sec.~\ref{sec:exp:agent}.}
    \label{fig:task_panels}
\end{figure}

\paragraph{Scene-Unaware Humanoid Motion Prior.}
We use a pretrained text-conditioned humanoid motion diffusion model~\cite{kimodo} as a differentiable motion prior, $M_\theta(\mathbf{z}\mid y,\mathcal{C})=\mathbf{x}$, mapping latent noise $\mathbf{z}$, a text prompt $y$, and optional sparse kinematic constraints $\mathcal{C}$ to a trajectory $\mathbf{x}$. It is trained on over 700 hours of human motion retargeted to humanoid embodiments, without explicit scene or interaction input. The prior can be conditioned on kinematic constraints (full-body keyframes, sparse joint positions/rotations, and dense root paths) passed to the model as inputs. But interaction requirements such as non-penetration, stable contact, and collision avoidance are \emph{implicit}: they depend on scene geometry and whole-body pose, cannot be written as keypoint targets in advance, and conditioning on these constraints alone does not yield contact-consistent motion (Sec.~\ref{sec:exp}).
\paragraph{Noise-Space Optimization.}
We optimize over the prior's latent variables rather than the 3D joint-position trajectory $\mathbf{x}$ directly, so every candidate remains a sample the prior can produce, retaining its learned realism and temporal coherence. Given latent noise $\mathbf{z} \in \mathbb{R}^{T \times D}$ unrolled through a differentiable DDIM sampling process, we solve
\begin{equation}
\label{eq:noiseopt}
\min_{\mathbf{z}}\;
\mathcal{L}\!\left(M_\theta(\mathbf{z} \mid y, \mathcal{C}),\, \mathcal{C}, \mathcal{S}\right)
=
w_g\,\mathcal{L}_{\mathrm{goal}}
+ w_c\,\mathcal{L}_{\mathrm{coll}}
+ w_f\,\mathcal{L}_{\mathrm{foot}}
+ w_h\,\mathcal{L}_{\mathrm{hand}}
+ w_e\,\mathcal{L}_{\mathrm{edge}},
\end{equation}
The goal term $\mathcal{L}_{\mathrm{goal}}$ is a masked loss on the constrained keypoints; the remaining terms are the scene-dependent collision, foot-contact, hand-contact, and edge-safety objectives; the last two are active only on tasks with grasps or terrain. We backpropagate through the entire denoising trajectory and optimize $\mathbf{z}$ with Adam; loss definitions and other details are in Appendix~\ref{app:opt}.
\paragraph{Behavior Diversity through Augmentations.}
For a fixed task specification, we generate diverse motions by randomizing initial poses, approach trajectories, and latent noise seeds (Figure~\ref{fig:task_panels}).

\subsection{Long-Horizon Multi-Stage Interaction Generation}
\label{sec:method:gen}
Single-window optimization is limited by the prior's temporal horizon and by GPU memory. We therefore decompose long behaviors into temporal windows $\{(y_w,\mathcal{C}_w)\}_{w=1}^{W}$ optimized in sequence: the decoded suffix of window $w$ is injected as a hard pose constraint into the prefix of window $w{+}1$ after re-canonicalization to the new heading and origin, composing long-horizon behavior without global optimization.

\subsection{Teacher-Student Visuomotor Learning}
\label{sec:method:teacherstudent}
\paragraph{Motion-Tracking Teacher.}
\label{sec:method:teacher}
For each task we train a privileged teacher on the $29$-DoF Unitree~G1 with PPO~\cite{schulman2017ppo}, through the training stack of~\cite{holosoma}, tracking the full batch of motions our framework generates for that task (Figure~\ref{fig:methodology}E). It observes proprioceptive state together with privileged reference and scene information and outputs $29$-D joint-angle residuals executed through PD control.
\paragraph{Depth-Conditioned Flow Student.}
\label{sec:method:student}
We distill the teacher into a student operating only from onboard observations: a history of head-mounted depth images and proprioception, encoded into a sensor context $\mathbf{c}$. Because the same context can be consistent with several valid interaction behaviors, we model the teacher's action distribution rather than its conditional mean, regressing a conditional velocity field with flow matching~\cite{lipman2023flowmatching,liu2023rectifiedflow}. Distillation is on-policy and at test time the field is integrated with a few fixed Euler steps; objective, architecture, and training details for both policies are in Appendix~\ref{app:policy}.

\section{Experiments}
\label{sec:exp}

We evaluate whether \model{} can generate humanoid--scene interaction motion references and whether those references provide effective supervision for humanoid control. Our experiments address four questions:
\textbf{(1)} Does noise-space optimization produce more contact- and scene-consistent motions than alternative methods of controlling the same prior (Sec.~\ref{sec:exp:motiongen})?
\textbf{(2)} Can the generated references be executed by physical policies and distilled to policies that use onboard sensing (Sec.~\ref{sec:exp:tracking})?
\textbf{(3)} Does tracking the generated references supervise better than task-based rewards, hierarchical control, or adversarial imitation (Sec.~\ref{sec:rltrain})?
\textbf{(4)} Can a coding agent reduce the engineering of extending \model{} to new interactions (Sec.~\ref{sec:exp:agent})?
We additionally present qualitative real-world deployments in videos available on the project page.

\begin{figure}[t]
    \centering
    \includegraphics[width=\linewidth]{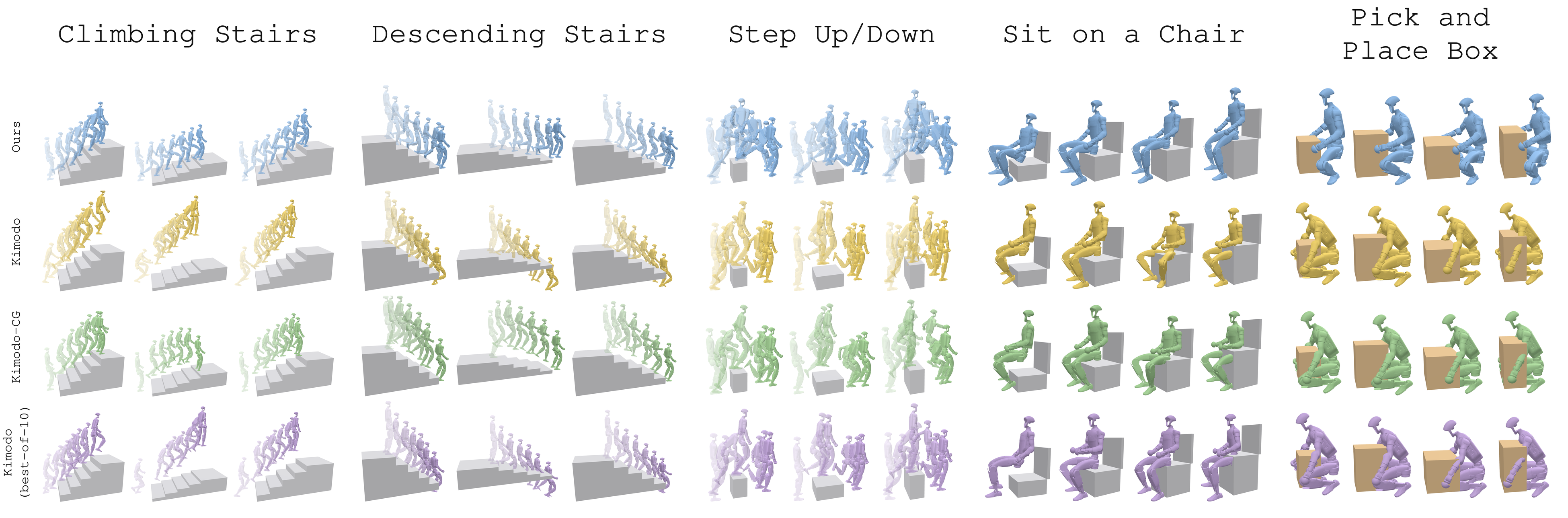}
\caption{\textbf{Humanoid--scene interaction synthesis with \model{} and baseline methods.} Rows, top to bottom: \model{} (ours, blue), kinematic-constraint conditioning (Kimodo, yellow), classifier guidance (Kimodo-CG, green), and best-of-ten sampling (Kimodo best-of-10, purple); each column shows the identical scene, spawn, and constraint set. Stair and step panels overlay time samples of one motion sequence (opacity increases with time); chair panels show the final seated frame, and box panels the frame after grasping.}
    \label{fig:data_gen_qual}
\end{figure}

\subsection{Humanoid--Scene Interaction Generation}
\label{sec:exp:motiongen}
\paragraph{Tasks.}
We evaluate motion generation on eight tasks spanning three forms of interaction (Table~\ref{tab:motiongen-detail}, Appendix~\ref{app:gen}): \textbf{free-space locomotion} (slalom walking among obstacles), \textbf{object interaction} (picking up, carrying, and placing boxes on the floor or a table), and \textbf{terrain interaction} (climbing and descending stairs, stepping onto and off blocks, sitting on and standing from chairs), randomizing scene geometry, object placement, interaction timing, and initial conditions throughout.

\paragraph{Baselines.}
We compare \model{} with three alternative mechanisms for controlling the same pretrained motion prior. All methods share prompts, random seeds, initial conditions, scene representation $\mathcal{S}$, and kinematic constraints $\mathcal{C}$. \textbf{Kimodo} samples the pretrained text-conditioned prior~\cite{kimodo} conditioned on the kinematic constraints $\mathcal{C}$, without access to the scene-dependent objectives defined by $\mathcal{S}$. \textbf{Kimodo-CG} receives the same constraints and scene information as \model{}, but applies the complete interaction objective as classifier guidance at each denoising step. \textbf{\model{}} instead optimizes the initial noise by backpropagating the objective through the complete denoising trajectory. \textbf{Kimodo best-of-10} draws ten independent samples per task specification and retains the one with the lowest interaction objective.

\paragraph{Metrics.}
We measure target satisfaction (\textbf{root path error}, \textbf{hand target error} at constrained frames) and geometric consistency (\textbf{scene penetration} over body query points, \textbf{foot--support gap} at the lower foot sole). The last two measure separate failure modes: a motion can close the gap by penetrating, or avoid penetration by floating. We report mean $\pm$ standard deviation over motion sequences.

\paragraph{Results.}
Per-task results are in Table~\ref{tab:motiongen-detail}, with full metric definitions in Appendix~\ref{app:gen}. \model{} achieves lower scene penetration across all three interaction categories, and lowers object-task hand-target error compared to the baselines.

Kimodo-CG is an informative comparison because it receives the same constraints, scene representation, and interaction objective as \model{}. Classifier guidance improves collision avoidance in free-space and terrain tasks but barely helps on object interaction, where dense root and hand conditioning leaves little room for the local corrections guidance can make.

Best-of-ten selection barely narrows the gap: matched sampling compute does not help, because the prior rarely samples a candidate satisfying every interaction requirement at once, which noise-space optimization instead constructs.

The box-pickup task shows this most clearly (Fig.~\ref{fig:data_gen_qual}). Across repeated samples and varying box sizes, the pretrained prior produces nearly identical wrist configurations and grasp widths, while \model{} adapts the arm trajectory, wrist configuration, and hand separation to each object's geometry.

\paragraph{Long-Horizon Multi-Stage Generation.} We apply the multi-stage procedure from Sec.~\ref{sec:method} to generate a four-stage pick-and-place sequence (Figure~\ref{fig:methodology}C): each stage is optimized with its own prompt and constraint set, with terminal poses transferred between consecutive windows.

\paragraph{Motion Repair.} The same optimization can repair imperfect retargeted human--humanoid interaction references when initialized from DDIM-inverted noise (Figure~\ref{fig:methodology}D); setup and qualitative results are in Appendix~\ref{app:repair}.

\subsection{Learning Interaction Humanoid Policies from Generated References} \label{sec:rltrain}

\paragraph{Do generated references supervise humanoid control better than task-level reinforcement-learning objectives?}
\model{} learns humanoid control by training privileged, scene-aware teacher policies to track its generated interaction motion references, which are then distilled into depth-conditioned students (Sec.~\ref{sec:exp:tracking}). Prior approaches instead optimize humanoid actions directly by reinforcement learning under task rewards, regularizing exploration with adversarial scene-unaware motion priors~\cite{peng2021amp} or building on pretrained low-level controllers~\cite{homie}. We compare these paradigms on \taskthree{} using a shared set of $20$ \model{}-generated motions.

\paragraph{Baselines.}
\begin{wrapfigure}{r}{0.46\textwidth}
\centering
\includegraphics[width=\linewidth]{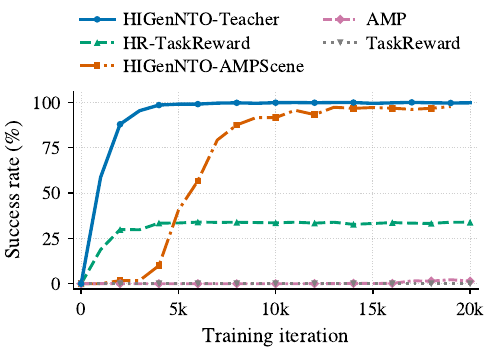}
\caption{\textbf{Policy learning on \taskthree{}.} Task success over reinforcement-learning iterations for generated reference-based motion tracking by \model{}-Teacher policies and four alternative training objectives. Each curve represents one training run.}
\label{fig:baseline_success}
\end{wrapfigure}
We compare \model{}-Teacher policies that track the complete generated trajectories, with four approaches:
\textbf{TaskReward} optimizes task rewards adapted from the Sit Down task of~\cite{physhsi}, with no motion prior, reference supervision, or contact and temporal supervision. \textbf{HR-TaskReward} trains a hierarchical policy under the same task rewards while using a frozen Homie controller~\cite{homie} as its low-level policy. \textbf{AMP}~\cite{peng2021amp} augments the task reward with an adversarial motion objective. Its discriminator is trained on the $20$ generated motion references \textit{excluding the scene}. \textbf{\model{}-AMPScene} additionally conditions the discriminator on chair-relative state from the generated references, providing interaction-aware adversarial supervision without trajectory tracking.

All methods use task goals and initial states derived from the same generated motions; AMP and \model{}-AMPScene additionally train their discriminators on them, and \model{}-Teacher tracks them directly.

\paragraph{Results.} Figure~\ref{fig:baseline_success} shows that the \model{}-Teacher policy reaches high task success in approximately one third of the training iterations required by \model{}-AMPScene, the strongest baseline. TaskReward fails to reliably discover the approach--contact--stabilization sequence from the task objective alone; HR-TaskReward and AMP add a pretrained controller and motion realism, respectively, but neither receives explicit supervision for the humanoid--chair interaction. Conditioning the discriminator on chair-relative state substantially improves \model{}-AMPScene, confirming that the generated references are also useful as a scene-aware motion prior; it still learns more slowly than \model{}-Teacher, whose tracking objective supplies a time-aligned target throughout the interaction.

The generated references thus carry temporal and geometric interaction structure that shaped task rewards and motion-only regularization do not recover. Fig.~\ref{fig:baseline_success} reports one run per method, and shows evidence of improved learning efficiency in this setting.

\paragraph{Can the generated interaction motions be executed without privileged information at test time?}
\label{sec:exp:tracking}

We compare \model{} privileged teacher and depth-based student policies in Table~\ref{tab:tracking-detail} (Appendix~\ref{app:track}). For context, we also evaluate SONIC~\cite{luo2026sonic}, a general-purpose motion tracker, zero-shot from its released weights: SONIC receives the motion reference but cannot observe the surrounding geometry, whereas the \model{} student observes the scene through onboard sensing but receives no motion reference, testing whether reference tracking alone suffices for interaction. All policies face the same eight tasks, scene configurations, and task-success criteria; the student, having no reference, is scored on task success only. Per-task results and complete metric definitions are provided there.

\paragraph{Results.} The privileged teachers execute the generated references reliably, and after distillation the students retain high success across all three interaction categories (Table~\ref{tab:tracking-detail}) while operating without privileged state or motion references; that interaction structure transfers to policies acting from onboard sensing alone. The student's long observation history matters on longer-horizon tasks: on \taskeight{}, the table is often occluded once the box is lifted, and the history lets the student retain the table position observed earlier in the rollout.

SONIC occasionally attains lower joint or root-orientation error, but its high root-position error and low task success show that scene-blind tracking reproduces plausible pose sequences without aligning them to the relevant object or support surface. 

\subsection{Scaling \model{} by Automating Task Specifications with a Coding Agent}
\label{sec:exp:agent}

Each task in Sec.~\ref{sec:method:formulation} is defined by a prompt $y$, constraints $\mathcal{C}$, and a scene representation $\mathcal{S}$; authoring and tuning these for every new interaction is a bottleneck. We investigate whether a coding agent can compile a natural-language task description into an executable \model{} specification.

\paragraph{Setup.}
Our implementation exposes a single task-level interface: every task is a function that takes a few randomization parameters (e.g.\ chair height, box size) and returns $(y,\mathcal{C},\mathcal{S})$, while the optimizer and the losses of Eq.~\ref{eq:noiseopt} stay fixed. Given a short brief and the prior's training-corpus annotations, the agent reads the corpus and writes a prompt $y$ framed in a similar way to the text used to train the motion prior, probes the frozen prior to size the scene $\mathcal{S}$ and constraints $\mathcal{C}$, drafts the task function, and renders a small batch to verify and revise it; a human only reviews the final motions with free-form feedback. The full procedure is detailed in Appendix~\ref{app:agenttasks}.

\paragraph{Results.}
Three of the eight tasks evaluated in Sec.~\ref{sec:exp:motiongen} (\tasknine{}, \taskten{}, and \taskeight{}) were specified through this procedure. In a more open-ended setting with no target task, the agent identifies capabilities represented in the corpus annotations and proposes compatible scene interactions; Figure~\ref{fig:task_panels} shows four resulting tasks (\taskeleven{}, \tasktwelve{}, \taskthirteen{}, \taskfourteen{}), spanning whole-body and non-prehensile object interaction with neither task-specific demonstrations nor manually authored constraints. Additional task definitions and qualitative results are provided in Appendix~\ref{app:agenttasks}.

\subsection{Real-World Deployment}
\label{sec:realworld}
We deploy depth-conditioned student policies for \taskthree{}, \taskfour{}, \taskeight{}, and \taskfive{} on a Unitree G1 using onboard depth and proprioception. Videos of the deployments and qualitative simulation results are available on the project page; Appendix~\ref{app:realworld} shows the rollouts.

\section{Conclusion and Limitations}

We presented \model{}, a framework for scalable humanoid interaction synthesis through constraint-guided latent optimization of pretrained motion priors. Rather than relying on large-scale interaction motion capture, \model{} generates interaction-rich humanoid motions from sparse differentiable objectives encoding contacts, geometry, and temporal interaction structure, and the resulting motions serve both as interaction motion references for humanoid control and as supervision for deployable visuomotor policies.
Our results suggest that interaction-rich behaviors need not be explicitly represented in the training distribution of a motion prior; they can emerge through optimization-time interaction constraints.

Our limitations suggest directions for future work. First, the generation objectives enforce kinematic and geometric consistency rather than dynamic feasibility; incorporating dynamics could directly produce motions satisfying balance, contact-force, and actuation constraints. Second, noise-space optimization remains computationally expensive and thus suited to offline data generation; flow-map-based optimization or amortized generation could reduce this cost. Finally, the framework addresses whole-body motion with simplified hand--object contact; dexterous manipulation will require richer hand representations, contact models, and object-dynamics objectives.

\bibliographystyle{plainnat}
\bibliography{bib/ref}

\clearpage
\appendix

\section{Real-world deployment}
\label{app:realworld}

\begin{figure}[t]
    \centering
    \includegraphics[width=\linewidth]{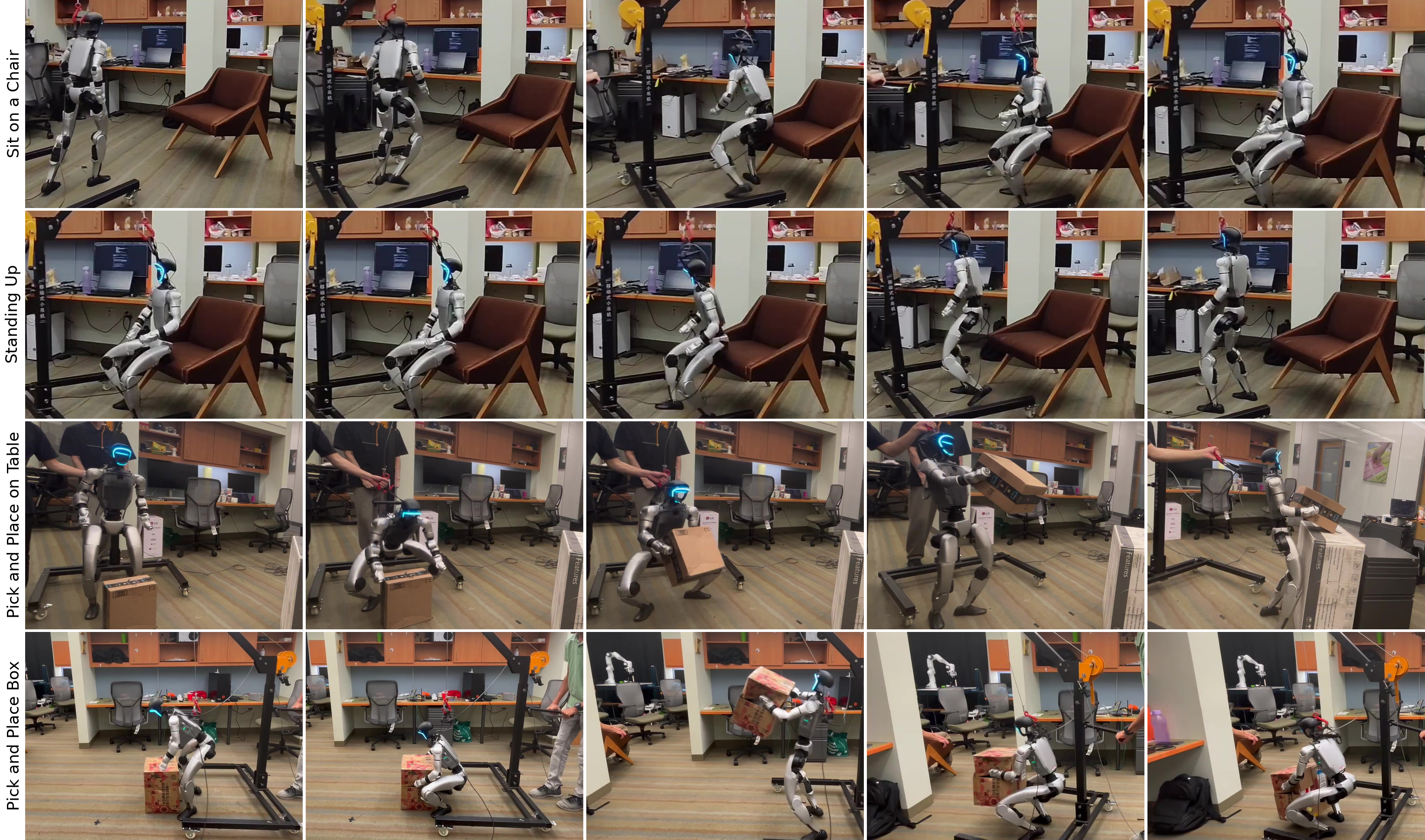}
    \caption{Real-world deployment of the distilled students, using only onboard sensor observations. The robot approaches a chair and sits down (\taskthree{}); stands back up from the seated state (\taskfour{}); lifts a box from the floor onto a table (\taskeight{}); and picks up and carries a box (\taskfive{}).}
    \label{fig:real_deploy}
\end{figure}

Figure~\ref{fig:real_deploy} shows the distilled student running on the physical robot for four tasks, from onboard sensing alone.

\section{Student evaluation in simulation}
\label{app:simeval}

\begin{figure}[!t]
    \centering
    \includegraphics[width=0.85\linewidth]{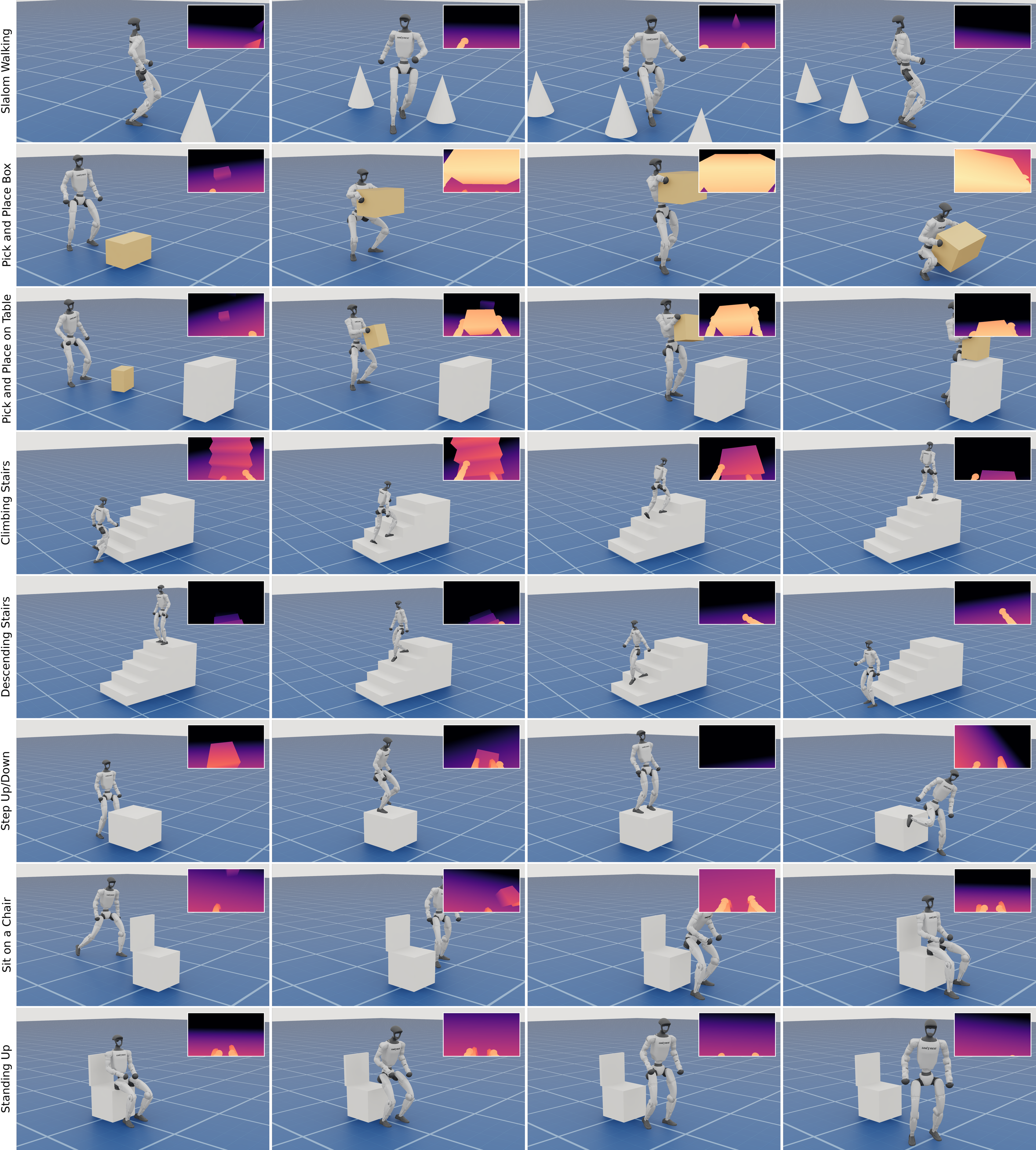}
    \caption{Evaluation of the trained student policies in simulation. The inset in each frame is the depth image the policy receives at that instant.}
    \label{fig:motion_generation_sim_deploy}
\end{figure}

Figure~\ref{fig:motion_generation_sim_deploy} shows the qualitative results for the trained students in simulation on all eight tasks, with depth and proprioception the policy's only inputs.

\section{Motion-generation results}
\label{app:gen}

\subsection{Per-task results}
\label{app:genresults}

\paragraph{Comparison setup.} Table~\ref{tab:motiongen-detail} reports the motion-generation results per task. All four methods are tested on an identical scene and constraint set. The baselines are sampled with $50$ DDIM denoising steps, while \model{} uses $10$ steps to limit the memory and compute required for backpropagation through the denoising process; Kimodo best-of-10 draws ten samples per motion sequence from the motion prior and keeps the one scoring lowest under the weighted objective of Appendix~\ref{app:losses}.

\paragraph{Runtime and compute.} One noise-optimization iteration of \model{} backpropagates through a $10$-step DDIM process (Alg.~\ref{alg:noiseopt}). On a single RTX A6000 it takes $9.7$--$26$\,s at batch size $20$ per iteration. The motions are optimized for $K{=}50$--$100$ iterations.

\begin{table}[t]
\centering
\scriptsize
\begin{tabular}{@{}lcccccccc@{}}
\toprule
& Free & \multicolumn{2}{c}{Object} & \multicolumn{5}{c}{Terrain} \\
\cmidrule(lr){2-2} \cmidrule(lr){3-4} \cmidrule(lr){5-9}
Method & Slalom  & Pick and  & Pick and Place & Climbing & Descending & Step    & Sit on  & Standing \\
       & Walking & Place Box & on Table       & Stairs   & Stairs     & Up/Down & a Chair & Up \\
\midrule
\multicolumn{9}{l}{\textit{Root path error (cm)} $\downarrow$} \\
\rowcolor{oursgreen}
Ours      & \textbf{\pmval{4.53}{0.82}} & \pmval{2.65}{0.26} & \pmval{2.73}{0.53} & \textbf{\pmval{2.78}{1.63}} & \textbf{\pmval{6.70}{3.40}} & \pmval{3.89}{0.61} & \textbf{\pmval{3.86}{0.74}} & \textbf{\pmval{2.87}{0.29}} \\
Kimodo    & \pmval{7.23}{3.20} & \pmval{2.93}{0.65} & \pmval{2.29}{0.51} & \pmval{34.00}{27.82} & \pmval{16.41}{7.50} & \pmval{2.86}{0.33} & \pmval{4.61}{1.15} & \pmval{2.96}{0.23} \\
Kimodo-CG & \pmval{7.23}{2.84} & \pmval{2.58}{0.32} & \textbf{\pmval{2.10}{0.35}} & \pmval{10.29}{8.85} & \pmval{12.81}{5.57} & \pmval{2.90}{0.66} & \pmval{4.49}{0.99} & \pmval{2.95}{0.26} \\
Kimodo best-of-10 & \pmval{7.22}{3.43} & \textbf{\pmval{2.48}{0.48}} & \pmval{2.30}{0.47} & \pmval{28.25}{19.41} & \pmval{17.06}{4.38} & \textbf{\pmval{2.76}{0.20}} & \pmval{4.46}{1.35} & \pmval{3.14}{0.19} \\
\midrule
\multicolumn{9}{l}{\textit{Hand target error (cm)} $\downarrow$} \\
\rowcolor{oursgreen}
Ours      & -- & \textbf{\pmval{6.36}{2.26}} & \textbf{\pmval{6.61}{2.67}} & -- & -- & -- & -- & -- \\
Kimodo    & -- & \pmval{18.93}{2.18} & \pmval{21.25}{1.78} & -- & -- & -- & -- & -- \\
Kimodo-CG & -- & \pmval{19.18}{2.23} & \pmval{20.60}{1.48} & -- & -- & -- & -- & -- \\
Kimodo best-of-10 & -- & \pmval{18.29}{2.37} & \pmval{19.97}{1.59} & -- & -- & -- & -- & -- \\
\midrule
\multicolumn{9}{l}{\textit{Scene penetration (cm)} $\downarrow$} \\
\rowcolor{oursgreen}
Ours      & \textbf{\pmval{0.01}{0.02}} & \textbf{\pmval{0.58}{0.46}} & \textbf{\pmval{0.51}{0.47}} & \pmval{1.10}{0.49} & \pmval{1.94}{0.62} & \textbf{\pmval{0.08}{0.08}} & \textbf{\pmval{1.25}{0.57}} & \textbf{\pmval{1.38}{1.63}} \\
Kimodo    & \pmval{33.96}{4.11} & \pmval{7.07}{6.74} & \pmval{10.35}{7.51} & \pmval{0.06}{0.08} & \pmval{232.79}{164.20} & \pmval{8.45}{3.39} & \pmval{9.42}{10.17} & \pmval{115.73}{118.03} \\
Kimodo-CG & \pmval{2.99}{0.95} & \pmval{7.95}{6.50} & \pmval{10.62}{7.09} & \textbf{\pmval{0.01}{0.02}} & \textbf{\pmval{1.63}{0.88}} & \pmval{1.97}{2.32} & \pmval{6.04}{7.22} & \pmval{107.59}{111.51} \\
Kimodo best-of-10 & \pmval{27.37}{2.84} & \pmval{4.99}{6.08} & \pmval{6.68}{5.70} & \pmval{0.04}{0.05} & \pmval{149.14}{103.53} & \pmval{4.63}{2.32} & \pmval{5.39}{6.26} & \pmval{86.75}{97.79} \\
\midrule
\multicolumn{9}{l}{\textit{Foot--support gap (cm)} $\downarrow$} \\
\rowcolor{oursgreen}
Ours      & \pmval{0.01}{0.04} & \textbf{\pmval{2.60}{0.22}} & \textbf{\pmval{2.82}{0.22}} & \textbf{\pmval{2.31}{0.30}} & \pmval{3.17}{1.21} & \textbf{\pmval{0.46}{0.59}} & \textbf{\pmval{0.21}{0.14}} & \textbf{\pmval{0.19}{0.10}} \\
Kimodo    & \textbf{\pmval{0.00}{0.01}} & \pmval{3.77}{0.80} & \pmval{3.24}{1.51} & \pmval{51.22}{14.42} & \textbf{\pmval{0.42}{0.27}} & \pmval{2.69}{2.06} & \pmval{0.75}{1.18} & \pmval{1.62}{2.93} \\
Kimodo-CG & \pmval{0.01}{0.06} & \pmval{3.17}{0.73} & \pmval{2.96}{0.56} & \pmval{24.57}{4.92} & \pmval{18.78}{5.65} & \pmval{3.48}{1.90} & \pmval{0.47}{0.24} & \pmval{1.66}{2.91} \\
Kimodo best-of-10 & \pmval{0.00}{0.00} & \pmval{5.25}{1.02} & \pmval{4.94}{1.27} & \pmval{39.46}{12.92} & \pmval{0.59}{0.30} & \pmval{2.99}{1.77} & \pmval{1.12}{1.21} & \pmval{1.92}{2.51} \\
\bottomrule
\end{tabular}
\caption{Motion-generation ablation, per task, on our dataset. All four methods are provided identical scene and constraints for each sequence. \textbf{Bold} $=$ best mean per metric.
Metric definitions in Appendix~\ref{app:genmetrics}.}
\label{tab:motiongen-detail}
\end{table}

\subsection{Evaluation metrics}
\label{app:genmetrics}

This section defines the four metrics of Table~\ref{tab:motiongen-detail}. All are reported in centimeters and lower is better. $\mathbf{x}^{t}_{j}\in\mathbb{R}^{3}$ is the world position of joint $j$ at frame $t$ of a generated motion sequence, over $T$ frames and $J$ skeleton joints. The contact threshold $\tau$ is the clearance within which a query point counts as touching a surface. The skin distance $\delta_j$ is the clearance each joint is expected to keep clear of static geometry. Against a moving obstacle the corresponding clearance is $\delta^{\mathrm{mov}}_{j}$.

\paragraph{Root path error.} The mean Euclidean distance, in the world ground plane, between the generated smoothed root position $\mathbf{r}^{t}$ and its target $\hat{\mathbf{r}}^{t}$, evaluated only at the constrained frames $\mathcal{T}_{r}$:
\begin{equation}
E_{\mathrm{root}} = \frac{1}{|\mathcal{T}_{r}|}\sum_{t\in\mathcal{T}_{r}} \big\| \Pi_{xy}\mathbf{r}^{t} - \hat{\mathbf{r}}^{t} \big\|_{2},
\end{equation}
with $\Pi_{xy}$ the projection onto the ground plane. Every task carries a root path constraint, so this is defined throughout; because that path is handed to all four methods, the metric reports how well each preserves a constraint it was given.

\paragraph{Hand target error.} The mean Euclidean distance between a constrained hand joint $j_e$ and its keyframe target $\hat{\mathbf{x}}^{t}_{j_e}$ over the specified keyframes $\mathcal{T}_{e}$:
\begin{equation}
E_{\mathrm{hand}} = \frac{1}{|\mathcal{T}_{e}|}\sum_{t\in\mathcal{T}_{e}} \big\| \mathbf{x}^{t}_{j_e} - \hat{\mathbf{x}}^{t}_{j_e} \big\|_{2}.
\end{equation}
It is defined only for tasks whose constraint set specifies hand end-effector targets, and reported as ``--'' elsewhere.

\paragraph{Scene penetration.} The per-frame penetration depth into scene geometry, summed over body query points and averaged over the sequence. Writing $s(\cdot)$ for the signed distance to the static scene and $s_{o}(\cdot)$ for the signed distance to moving obstacle $o$,
\begin{equation}
\begin{aligned}
E_{\mathrm{pen}} = \frac{1}{T}\sum_{t=1}^{T} \Big[ &\textstyle\sum_{j}\max\!\big(0,\ \delta_j - s(\mathbf{x}^{t}_{j})\big) \\
+ &\textstyle\sum_{p\in\mathcal{P}}\max\!\big(0,\ \tau - s(\mathbf{x}^{t}_{p})\big) \\
+ &\textstyle\sum_{o}\sum_{j}\max\!\big(0,\ \delta^{\mathrm{mov}}_{j} - s_{o}(\mathbf{x}^{t}_{j})\big) \Big],
\end{aligned}
\end{equation}
where $\mathcal{P}$ is the set of sole points of both feet; querying the sole rather than the ankle joint captures the actual surface of the feet. Joints gripping a moving obstacle are excluded while the object is held, so an intended grasp is not penalized, but other joints are penalized for penetrating the obstacle.

\paragraph{Foot--support gap.} The mean height of the lower foot's nearest sole point above the support surface, beyond a contact threshold:
\begin{equation}
E_{\mathrm{gap}} = \frac{1}{T}\sum_{t=1}^{T} \max\!\Big(0,\ \min_{p\in\mathcal{P}} s(\mathbf{x}^{t}_{p}) \;-\; \tau\Big).
\end{equation}
The minimum runs over the same $\mathcal{P}$, so it selects the lowest sole point of whichever foot is lower.

\paragraph{Why the last two are reported together.} Scene penetration and foot--support gap are the two one-sided halves of a single signed clearance: a motion can drive penetration to zero by hovering above the scene entirely, and can close the support gap by penetrating through it. Both failure modes occur among the baselines (Table~\ref{tab:motiongen-detail}), which is why we report the pair.

\section{Repairing retargeted interaction motions}
\label{app:repair}

\begin{wrapfigure}{r}{0.5\textwidth}
    \centering
    \includegraphics[width=\linewidth]{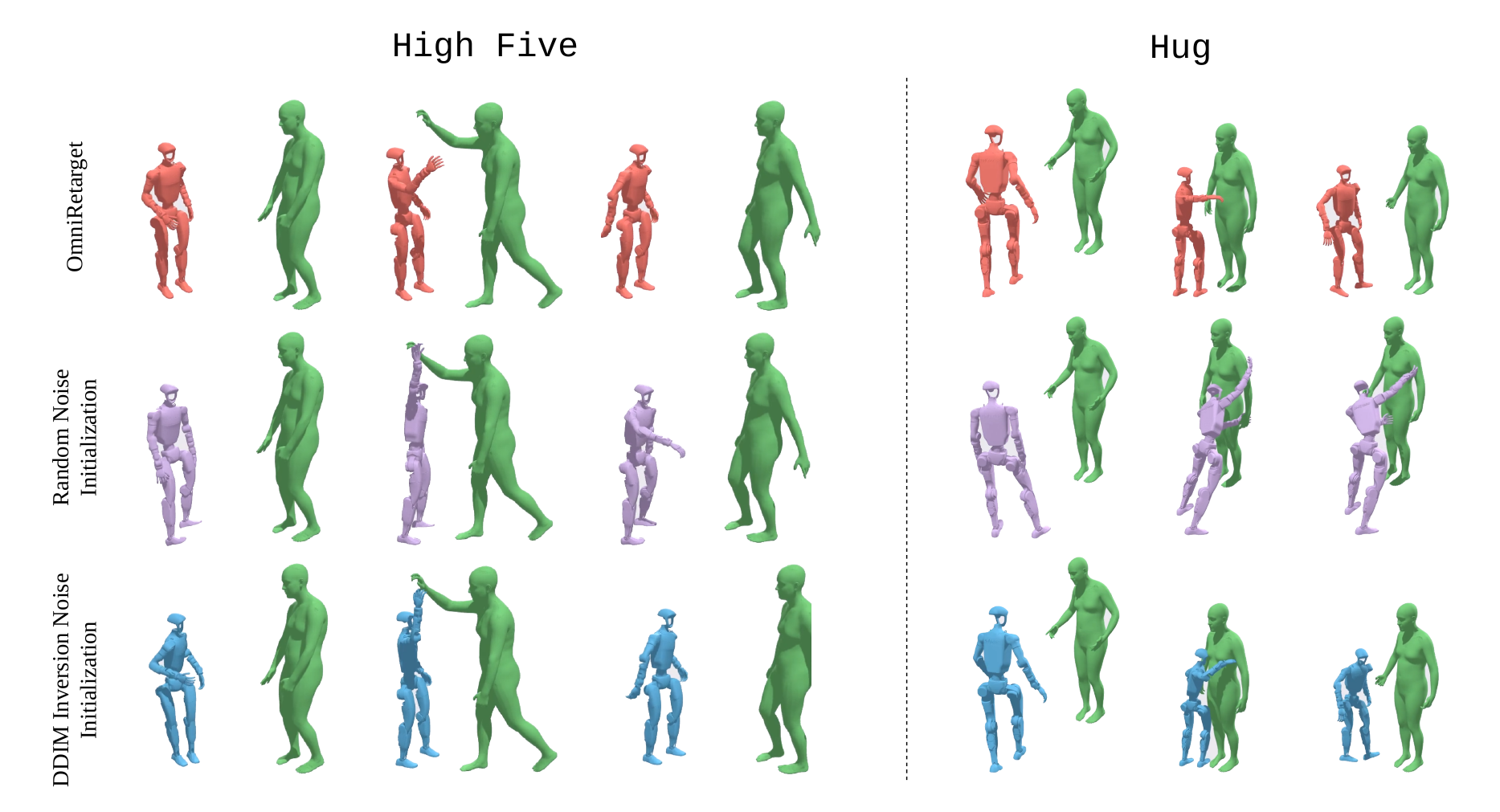}
    \caption{\textbf{Motion reference repair.} \model{} initialized with DDIM-inverted noise.}
    \label{fig:ddim_inv_refine}
\end{wrapfigure}

The same optimization framework can repair an existing interaction trajectory by initializing from its inverted latent noise rather than from random noise. We demonstrate this on two Inter-X sequences~\cite{interx}, \tasksix{} and \taskseven{}, retargeted with OmniRetarget~\cite{yang2025omniretarget}: in the resulting references the hands fail to meet during the high-five, while the hug shows body interpenetration without the intended surface contact (Figure~\ref{fig:ddim_inv_refine}, first row).

To construct the repair constraints we threshold inter-body distances in the original human--human sequence for candidate contact locations and times, then use a vision-language model to retain the interaction-relevant ones as $\mathcal{C}$. The human is represented by the SMPL-X~\cite{smplx2019} body SDF from VolumetricSMPL~\cite{volsmpl}, and the latent is initialized by DDIM inversion~\cite{ddim} of the retargeted trajectory.

\model{} then restores the intended contacts in both interactions while preserving the structure of the original motion (Figure~\ref{fig:ddim_inv_refine}, last row). Random initialization suffices for the simpler high-five but fails on the more tightly constrained hug (second row), indicating that inversion provides an important initialization when repairing interactions with dense contact requirements.

\section{Motion-tracking results}
\label{app:track}

\subsection{Per-task results}
\label{app:trackresults}

Table~\ref{tab:tracking-detail} reports the motion-tracking results per task, aggregated by interaction type (free-space, object, terrain) in the main text. The two trackers are scored on every metric and the student, which receives no motion reference, is scored on task success alone.

\begin{table}[t]
\centering
\scriptsize
\begin{tabular}{@{}lcccccccc@{}}
\toprule
& Free & \multicolumn{2}{c}{Object} & \multicolumn{5}{c}{Terrain} \\
\cmidrule(lr){2-2} \cmidrule(lr){3-4} \cmidrule(lr){5-9}
Method & Slalom  & Pick and  & Pick and Place & Climbing & Descending & Step    & Sit on  & Standing \\
       & Walking & Place Box & on Table       & Stairs   & Stairs     & Up/Down & a Chair & Up \\
\midrule
\multicolumn{9}{l}{\textit{Joint error (rad)} $\downarrow$} \\
\rowcolor{oursgreen}
Teacher (ours)  & \textbf{\pmval{0.141}{0.014}} & \pmval{0.182}{0.022} & \pmval{0.296}{0.065} & \textbf{\pmval{0.087}{0.008}} & \textbf{\pmval{0.107}{0.011}} & \textbf{\pmval{0.099}{0.010}} & \textbf{\pmval{0.087}{0.008}} & \textbf{\pmval{0.068}{0.012}} \\
SONIC           & \pmval{0.231}{0.059} & \textbf{\pmval{0.114}{0.009}} & \textbf{\pmval{0.100}{0.025}} & \pmval{0.326}{0.128} & \pmval{0.220}{0.104} & \pmval{0.124}{0.050} & \pmval{0.115}{0.025} & \pmval{0.388}{0.138} \\
\midrule
\multicolumn{9}{l}{\textit{Root position error (m)} $\downarrow$} \\
\rowcolor{oursgreen}
Teacher (ours)  & \textbf{\pmval{0.131}{0.047}} & \textbf{\pmval{0.076}{0.010}} & \textbf{\pmval{0.128}{0.069}} & \textbf{\pmval{0.056}{0.008}} & \textbf{\pmval{0.060}{0.009}} & \textbf{\pmval{0.059}{0.007}} & \textbf{\pmval{0.058}{0.008}} & \textbf{\pmval{0.062}{0.016}} \\
SONIC           & \pmval{1.153}{0.387} & \pmval{0.462}{0.080} & \pmval{0.526}{0.151} & \pmval{1.399}{0.297} & \pmval{0.763}{0.278} & \pmval{1.538}{0.670} & \pmval{0.341}{0.124} & \pmval{1.006}{0.174} \\
\midrule
\multicolumn{9}{l}{\textit{Root orientation error (rad)} $\downarrow$} \\
\rowcolor{oursgreen}
Teacher (ours)  & \textbf{\pmval{0.221}{0.047}} & \pmval{0.193}{0.033} & \pmval{0.301}{0.119} & \textbf{\pmval{0.144}{0.021}} & \textbf{\pmval{0.168}{0.031}} & \textbf{\pmval{0.127}{0.022}} & \textbf{\pmval{0.145}{0.020}} & \textbf{\pmval{0.140}{0.034}} \\
SONIC           & \pmval{0.453}{0.167} & \textbf{\pmval{0.153}{0.009}} & \textbf{\pmval{0.154}{0.060}} & \pmval{0.747}{0.324} & \pmval{0.485}{0.269} & \pmval{0.224}{0.107} & \pmval{0.198}{0.041} & \pmval{0.864}{0.368} \\
\midrule
\multicolumn{9}{l}{\textit{Task success (\%)} $\uparrow$} \\
\rowcolor{oursgreen}
Teacher (ours)  & \textbf{\pmval{97.7}{2.1}} & \textbf{\pmval{99.5}{0.8}} & \textbf{\pmval{71.1}{34.8}} & \textbf{\pmval{99.7}{0.2}} & \textbf{\pmval{99.5}{0.5}} & \textbf{\pmval{100.0}{0.0}} & \textbf{\pmval{99.9}{0.1}} & \textbf{\pmval{97.6}{2.2}} \\
\rowcolor{oursgreen}
Student (ours)  & \pmval{65.9}{25.4} & \pmval{80.9}{18.0} & \pmval{57.4}{36.7} & \pmval{96.5}{5.2} & \pmval{95.2}{4.9} & \pmval{98.3}{1.6} & \pmval{88.5}{20.1} & \pmval{96.3}{4.2} \\
SONIC           & \pmval{3.1}{4.9} & \pmval{8.0}{8.3} & \pmval{0.0}{0.0} & \pmval{0.2}{0.5} & \pmval{28.8}{23.0} & \pmval{0.2}{0.4} & \pmval{19.5}{22.6} & \pmval{3.9}{7.0} \\
\bottomrule
\end{tabular}
\caption{Motion tracking, per task. \textbf{Bold} $=$ best mean per metric. Setup and rubrics in Appendix~\ref{app:trackmetrics}.}
\label{tab:tracking-detail}
\end{table}

\subsection{Evaluation metrics and setup}
\label{app:trackmetrics}

\paragraph{Metrics.} The three error metrics are accumulated once per control step and averaged over the attempt. \emph{Joint error} is the mean absolute difference between reference and measured joint angles over the $29$ actuated degrees of freedom, in radians. \emph{Root position error} is the Euclidean distance between reference and measured root positions in the world frame, in meters, and \emph{root orientation error} the geodesic angle between the two root quaternions, in radians.

\paragraph{Uncensored attempts.} Every attempt plays its reference from the first frame to the last: terminations are recorded, but the environment keeps running rather than resetting. This avoids the bias of censored measurements, under which a policy that terminates early is scored almost entirely on the easy opening of every sequence.

\paragraph{Task success.} \emph{Task success} measures how well the expected behavior is executed. It combines a small set of criteria selected per task, derived from the reference motions: terminal root height and uprightness for stair climbing; the same plus terminal root location for stair descent and standing up; terminal root location and height for sitting; terminal root height, uprightness, and peak height for stepping onto a box; terminal root location and uprightness for slalom walking; and terminal object position for the box-manipulation tasks. 

\paragraph{Evaluation setup.} Each policy is rolled with $4096$ environments for the teacher and SONIC and $384$ for the students, which are limited by the memory requirements of the simulated depth camera. Students are evaluated on clean rendered depth with no augmentation.

\section{Task specifications}
\label{app:tasks}

\subsection{Per-task specifications}
\label{app:taskspec}

Every task is specified through the three components of Sec.~\ref{sec:method:formulation}: a prompt $y$ for the frozen Kimodo prior~\cite{kimodo}, sparse spatiotemporal constraints $\mathcal{C}$ on a few keypoints (typically root/pelvis and hands, plus a heading direction), and a scene $\mathcal{S}$ of SDFs and terrain heightfields. It needs no additional data collection. The feet are never given targets: foot placement comes entirely from the contact and edge losses against the scene. The chair, object, and slalom tasks sample an initial root pose and an obstacle-avoiding path, which is their main source of behavioral diversity; the stair and step tasks prescribe the root trajectory from the geometry instead, varying the terrain and the sampled noise. Per-task loss weights are in Appendix~\ref{app:losses}.

\begin{itemize}
\item \taskone{}: prompt ``A person climbs up stairs.'' The root follows a stepped path up the staircase while the foot-contact and edge losses place the feet on the treads. \emph{Scene:} a closed-form staircase SDF over a ground plane, with a heightfield obtained from the same geometry, which is used to compute the edges for $\mathcal{L}_{\mathrm{edge}}$. \emph{Randomization:} tread (step depth) $\in\{0.3,0.4\}$\,m $\times$ rise (step height) $\in\{0.1,0.2\}$\,m over five steps.

\item \tasktwo{}: prompt ``A person climbs down stairs.'' \emph{Scene} and \emph{randomization} as for \taskone{}.

\item \taskten{}: prompts ``A person climbs up a box.'' then ``A person climbs down a box.'', optimized as two overlapping windows. The root follows a prescribed trajectory: a walk in, a rise onto the box, a pause on top, and a drop. \emph{Scene:} a closed-form box SDF over a ground plane. \emph{Randomization:} box heights $\{0.3,0.4,0.5\}$\,m against depths $\{0.3,0.6\}$\,m.

\item \taskthree{}: prompt ``A person walks for \mbox{sometime} and sits down on a chair.'' The root follows an approach path and is then held at the seat, with the robot turned to face away from the backrest, as it must be to sit. \emph{Scene:} a parametric chair SDF consisting of $2$ cuboids (seat and backrest) on a ground plane. \emph{Randomization:} seat height $\in\{0.2,0.3,0.4,0.5\}$\,m, with start positions drawn within a $2.0$\,m radius.

\item \taskfour{}: prompt ``A person sitting on a chair stands up and walks forward.'', which starts the motion seated without any explicit initialization. The root is held at the seat, with its heading held in the same seated orientation as \taskthree{}; it is then constrained along a prescribed path $0.5$\,m out from the chair's front edge. There is no explicit stand-up constraint, the rise results from the prompt. \emph{Scene:} the same chair SDF as \taskthree{}. \emph{Randomization:} seat height $\in\{0.2,0.3,0.4,0.5\}$\,m.

\item \tasknine{}: prompt ``A person walks forward, turning left and right to avoid obstacles.'' A path weaving between the pillars is prescribed and paced at $1.0\,\mathrm{m/s}$; the root tracks it and the pelvis height is held at its nominal standing value. \emph{Scene:} a union of SDFs of pillars which act as obstacles over a ground plane. \emph{Randomization:} pillar shape $\in\{\text{box},\text{cone}\}$ $\times$ count $\in\{3,4,5\}$ $\times$ spacing $\in\{0.9,1.2,1.5\}$\,m.

\item \taskfive{}: prompts ``A person walks.'', ``A person lifts a crate with both hands.'', ``A person carrying a crate with both hands walks forward.'', and ``A person places a crate on the floor with both hands.'', optimized as four overlapping windows. Both hand end effectors are constrained to the box's side faces and the pelvis height profile is prescribed for the lift and the set-down. \emph{Scene:} a ground plane and a moving box SDF driven by an object trajectory. \emph{Randomization:} boxes with side lengths in $[0.15,0.50]$\,m; the start pose is drawn from a wedge-shaped region $0.55$--$1.0$\,m behind the box.

\item \taskeight{}: the same four windows as \taskfive{}, with the fourth prompt ``A person puts a crate down with both hands.'' Only the destination for the constraints differs: the hands settle at the table's height and the pelvis height is constrained to remain near standing height rather than dropping into a squat. \emph{Scene:} adds a static table cuboid. \emph{Randomization:} boxes with side lengths in $[0.15,0.50]$\,m, with table height $\in\{0.2,0.4,0.6\}$\,m and table width $\in\{0.3,0.5\}$\,m at a fixed $0.6$\,m length.
\end{itemize}

\subsection{Agent-authored task specification}
\label{app:agenttasks}

\begin{figure}[t]
    \centering
    \includegraphics[width=\linewidth]{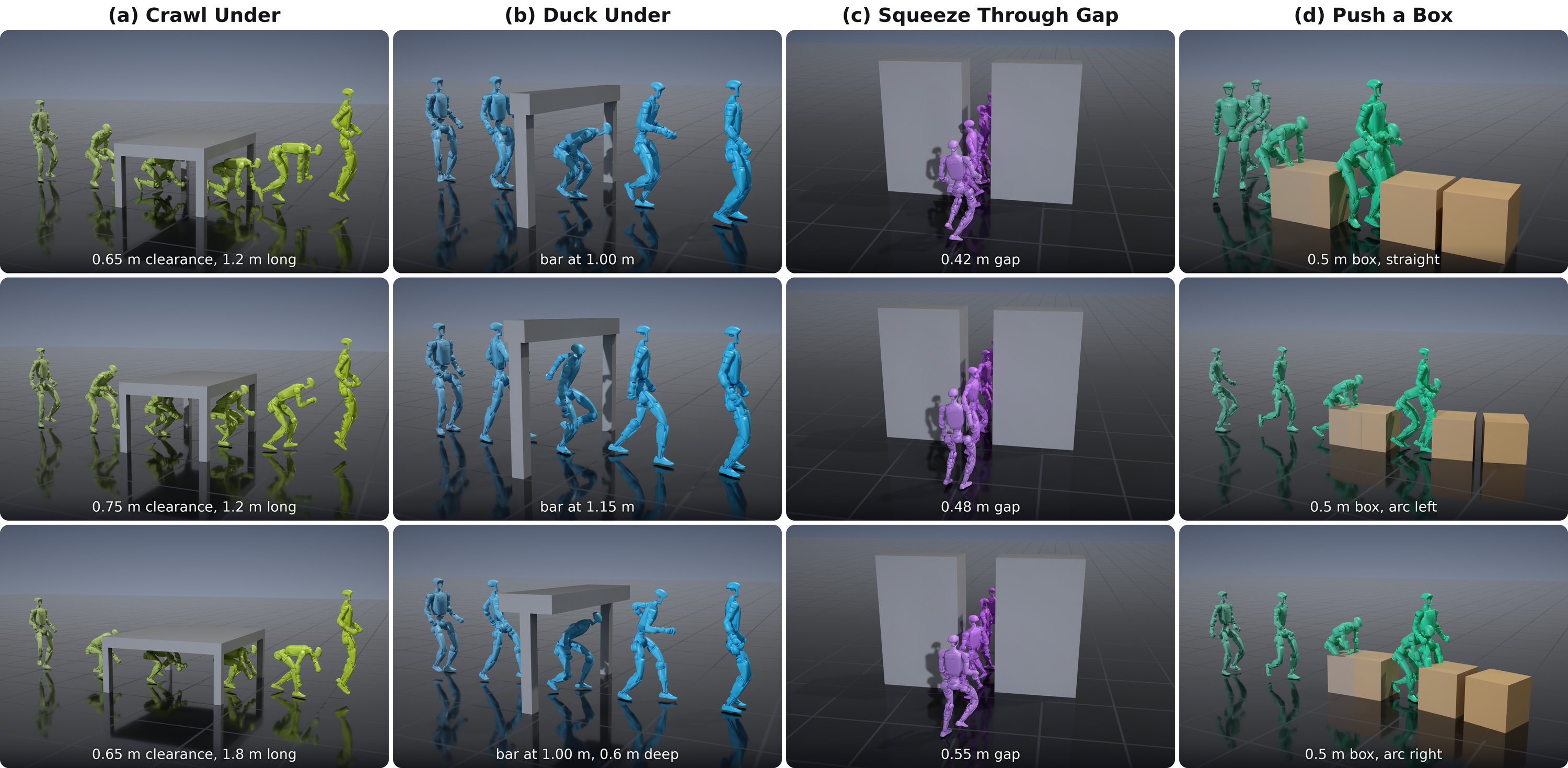}
    \caption{\textbf{Agent-authored behaviors across their randomization settings.} Each panel overlays frames from a single generated motion. The three rows are different scene configurations, and the columns are (a) \taskeleven{}, varying slab clearance and length; (b) \tasktwelve{}, varying slab clearance and depth; (c) \taskthirteen{}, varying gap width; and (d) \taskfourteen{}, varying box path. The prompt and constraints behind every panel were written by the coding agent, with no task-specific demonstrations and no hand-authored constraints.}
    \label{fig:agent_task_panels}
\end{figure}

This section details the procedure of Sec.~\ref{sec:exp:agent}, by which a frontier coding agent (Claude Code running Claude Fable 5) authors the task programs consumed by \model{}.

\paragraph{Task interface.} The agent writes a single function per task. This function takes a few task-specific parameters as input and returns the prompt $y$, the constraints $\mathcal{C}$, and the scene $\mathcal{S}$ of Sec.~\ref{sec:method:formulation}, together with the object trajectory and the frames in which contact is required. Varying those parameters is what yields the diverse scene configurations and initial poses for each task. The agent is instructed not to modify the optimizer or the loss terms.

\paragraph{Procedure.} Given a short natural-language brief and a pointer to the text annotations of the corpus the prior was trained on, the agent: \emph{(i)}~mines the annotations for the group describing the requested behavior and adopts the corpus's own phrasing as the prompt $y$; \emph{(ii)}~probes the frozen prior by running the candidate prompts with no scene and no constraints, then measures the resulting motion for an initial estimate of plausible scene sizes and constraints; \emph{(iii)}~drafts the first iteration of the function that builds the scene and constraints for the task; \emph{(iv)}~generates a batch of motions ($2$ settings $\times$ $5$ motions $\times$ $100$ optimization steps), reports per-term losses, renders the motions as videos to check that they follow the intended behavior and look natural, and iterates if needed. Steps \emph{(ii)} and \emph{(iv)} close the loop: the agent verifies that a task works before involving a human.

\paragraph{Agent-authored behaviors.} An agent can author tasks in two ways. In the \emph{user-specified} mode the human names the behavior (\taskeight{}, \tasknine{}, \taskten{}). In the \emph{discovery} mode (\taskeleven{}, \tasktwelve{}, \taskthirteen{}, \taskfourteen{}) the human names no task at all: given only the corpus annotations, the agent surveys the prior's vocabulary and returns grounded task proposals, each citing the annotation group that makes it in-distribution, sketching an implementation, and proposing axes along which to randomize the motions. We use the annotations from the BONES-SEED dataset~\cite{bones-seed-2026}, used in Kimodo~\cite{kimodo}.

Figure~\ref{fig:agent_task_panels} shows generated motions for each of these discovered tasks, with three scene configurations per task. All four use the same interface as the tasks of Appendix~\ref{app:taskspec}, and their optimizer settings and loss weights are listed alongside them in Table~\ref{tab:lossweights}.

\begin{itemize}
\item \taskeleven{}: prompts ``A person walks forward and crouches down.'', ``A person crawls forward on hands and knees.'', and ``A person stands up and walks forward.'', generated over three windows. The root follows a prescribed path whose height descends from standing to a hands-and-knees pelvis height (this height is found by probing the prior as described in step \emph{(ii)}) before the slab's near face, is held through the slab span, and rises only once the far face is cleared. Each window boundary therefore falls where the body is momentarily at rest. \emph{Scene:} a horizontal slab raised on four corner posts over a ground plane ($5$ cuboid SDFs), where the \emph{clearance} is the free gap between the ground and the slab's underside. \emph{Randomization:} clearance $\in\{0.65,0.75\}$\,m and slab length $\in\{1.2,1.8\}$\,m, with start positions randomized up to $0.25$\,m back along the approach path.

\item \tasktwelve{}: prompt ``A person walks forward and bends down to avoid an obstacle.'' The root follows a straight path at a constant speed, with a pelvis-height constraint that dips across the slab span and returns to standing once past it; the SDF-based loss against the slab forces the head to clear it. \emph{Scene:} an overhead slab spanning the path on two side posts over a ground plane ($3$ cuboid SDFs), where the \emph{clearance} is the gap between the ground and the slab's underside. \emph{Randomization:} clearance $\in\{1.00,1.15\}$\,m and slab depth $\in\{0.20,0.60\}$\,m, with start positions randomized up to $0.30$\,m back along the approach path.

\item \taskthirteen{}: prompt ``A person turns sideways and walks through a narrow gap.'' The root path runs straight through the gap center, while a $\pm\pi/2$ heading constraint on the frames at the narrowest point turns the body sideways for the squeeze. \emph{Scene:} two facing wall blocks ($2$ cuboid SDFs, $2.0$\,m tall) on a ground plane, leaving a corridor of \emph{gap width} across the path. \emph{Randomization:} gap width $\in\{0.42,0.48,0.55\}$\,m, the leading shoulder sampled left or right per motion sequence, and start positions randomized up to $0.25$\,m back along the approach path.

\item \taskfourteen{}: prompts ``A person walks.'', ``A person strains while slowly moving forward, pushing a large box.'', and ``A person stands still.'', generated over three windows. The path of the box is prescribed first as a constant-curvature arc, and both hand targets are derived from it as rigid offsets on the box's rear face. The root is constrained to trail the rear face at a fixed distance, with its heading held to the path tangent and its pelvis lowered through the pushing phase; it steps back once the box is released. \emph{Scene:} a ground plane and a rigid box represented as a moving box SDF following the prescribed object trajectory. \emph{Randomization:} box edge $\in\{0.5,0.6\}$\,m $\times$ path $\in\{$straight, left arc, right arc$\}$ with curvature $\pm0.15\,\mathrm{m}^{-1}$, pushed $1.75$--$2.0$\,m, with start positions drawn from a wedge-shaped region $0.9$--$1.8$\,m behind the box.
\end{itemize}

\section{Constraint-guided motion optimization}
\label{app:opt}

\subsection{Optimization algorithm}
\label{app:algo}

\begin{algorithm}[t]
\caption{Constraint-guided motion optimization.}
\label{alg:noiseopt}
\begin{algorithmic}[1]
\Require prompt $y$, constraints $\mathcal{C}$, scene $\mathcal{S}$, outer steps $K$, DDIM steps $N$, batch size $B$
\State Initialize $\mathbf{z} \sim \mathcal{N}(0, \sigma_{\mathrm{init}}^2\mathbf{I})$ for $B$ samples
\For{$k = 1, \dots, K$}
  \State $\mathbf{x} \gets M_\theta(\mathbf{z} \mid y, \mathcal{C})$ \Comment{full $N$-step DDIM denoise with gradients enabled}
  \State Align $\mathbf{x}$ to the global coordinate frame at the desired frame-0 root position and orientation
  \State $\mathcal{L} \gets w_g \mathcal{L}_{\mathrm{goal}}(\mathbf{x}, \mathcal{C}) + w_c \mathcal{L}_{\mathrm{coll}}(\mathbf{x}, \mathcal{S}) + w_f \mathcal{L}_{\mathrm{foot}}(\mathbf{x}, \mathcal{S}) + w_h \mathcal{L}_{\mathrm{hand}}(\mathbf{x}, \mathcal{S}) + w_e \mathcal{L}_{\mathrm{edge}}(\mathbf{x}, \mathcal{S})$ \label{line:loss}
  \State $\mathbf{g} \gets \nabla_{\mathbf{z}} \mathcal{L}$
  \State $\mathbf{z} \gets \textsc{Adam}(\mathbf{z}, \mathbf{g})$;\;\; record per-sample best $\mathbf{x}^\star_b$ at the lowest $\mathcal{L}_b$ seen so far
\EndFor
\State \Return $\{\mathbf{x}^\star_b\}_{b=1}^{B}$
\end{algorithmic}
\end{algorithm}

\subsection{Loss terms}
\label{app:losses}

We detail the loss terms summed in line~\ref{line:loss} of Alg.~\ref{alg:noiseopt}. Let $\mathbf{x}^{t}_{j}\in\mathbb{R}^3$ denote the world-frame position of joint $j$ at frame $t$ for a single sample. The scene SDF $s(\cdot)$ (negative inside geometry), the contact threshold $\tau$, and the skin distances $\delta_j$ and $\delta^{\mathrm{mov}}_{j}$ are as defined in Appendix~\ref{app:genmetrics}, as is the set of sole points $\mathcal{P}$; $\mathcal{F}=\{\text{left},\,\text{right}\}$ indexes the two feet.

\paragraph{Goal ($\mathcal{L}_{\mathrm{goal}}$).} A mask-normalized Huber loss between the denoised joints and the targets $\hat{\mathbf{x}}^{t}_{j}$ obtained from $\mathcal{C}$, gated by a per-frame mask $m^{t}_{j}\in\{0,1\}$ that selects only the constrained $(t,j)$ entries:
\begin{equation}
\mathcal{L}_{\mathrm{goal}} = \frac{\sum_{t,j} m^{t}_{j}\, H_{1}\!\big(\mathbf{x}^{t}_{j},\, \hat{\mathbf{x}}^{t}_{j}\big)}{\sum_{t,j} m^{t}_{j}},
\end{equation}
where $H_{1}$ is the Huber loss ($\delta_H=1$) summed over the three coordinates. This single term handles the constraints on the root, hand, and full-body; the mask is the only thing that changes across tasks.

\paragraph{Foot contact ($\mathcal{L}_{\mathrm{foot}}$).} Penalizes the foot nearest to the ground from floating, encouraging at least one foot to stay in contact at all times. It is queried at points on the sole of the feet $\mathcal{P}$.
\begin{equation}
\mathcal{L}_{\mathrm{foot}} = \frac{1}{T}\sum_{t=1}^{T} \max\!\Big(0,\; \min_{p\in\mathcal{P}} s(\mathbf{x}^{t}_{p}) - \tau\Big).
\end{equation}

\paragraph{Hand contact ($\mathcal{L}_{\mathrm{hand}}$).} Holds the hands at a fixed palm clearance $d_{\mathrm{palm}}$ off the object's faces, measured against the object's moving SDF over the grip window $\mathcal{G}$ and the hand keypoints $\mathcal{H}$. Unlike the terms above it is two-sided (a hand drifting off the object is penalized as much as one sinking into it) and tolerates a deviation of up to $\tau$ from the target clearance:
\begin{equation}
\mathcal{L}_{\mathrm{hand}} = \frac{1}{|\mathcal{G}|\,|\mathcal{H}|}\sum_{t\in\mathcal{G}}\sum_{j\in\mathcal{H}} \max\!\Big(0,\ \big| s_{o}(\mathbf{x}^{t}_{j}) - d_{\mathrm{palm}} \big| - \tau\Big).
\end{equation}
It is active on the tasks involving object manipulation.

\paragraph{Collision ($\mathcal{L}_{\mathrm{coll}}$).} Penalizes penetration of any joint into scene geometry up to its skin distance:
\begin{equation}
\mathcal{L}_{\mathrm{coll}} = \frac{1}{T}\sum_{t=1}^{T}\sum_{j=1}^{J} \max\!\big(0,\; \delta_j - s(\mathbf{x}^{t}_{j})\big).
\end{equation}
Each moving obstacle contributes an additional term of the same form, queried against its per-frame (moving) SDF.

\paragraph{Edge contact ($\mathcal{L}_{\mathrm{edge}}$).} This term keeps the planted foot away from sharp terrain edges. Each foot $f$ carries a query set $\mathcal{Q}_f$ of its ankle and toe points, and we define a per-point risk as a linear ramp in the distance $d(q)$ to the nearest edge,
\begin{equation}
\ell(q) = 1 - \mathrm{clamp}\!\big(d(q)/\rho,\,0,\,1\big) \in [0,1],
\end{equation}
where $\rho$ is a safety radius and $d(q)$ is the distance from an edge precomputed from the terrain heightfield. The planted foot $f^\star$ is the one whose lowest query point sits closest to the terrain,
\begin{equation}
f^\star = \arg\min_{f\in\mathcal{F}} \; \min_{q\in\mathcal{Q}_f} \big(z(q^{t}) - h(q^{t})\big),
\end{equation}
with $z(\cdot)$ the height of a query point and $h(\cdot)$ the terrain height beneath it. The term is the planted foot's mean point risk, averaged over the sequence,
\begin{equation}
\mathcal{L}_{\mathrm{edge}} = \frac{1}{T}\sum_{t=1}^{T} \frac{1}{|\mathcal{Q}_{f^\star}|}\sum_{q\in\mathcal{Q}_{f^\star}} \ell\!\big(q^{t}\big).
\end{equation}
The choice of $f^\star$ is treated as a constant in the backward pass, so gradients flow through the selected foot's risk but not through which foot is selected.

\paragraph{Loss weights.} The goal weight is $w_g{=}1.0$ throughout; the remaining weights and the optimizer settings are per task, and are listed in Table~\ref{tab:lossweights}. The shared constants are the contact threshold $\tau{=}0.01\,\mathrm{m}$; the skin distance $\delta_j{=}0.05\,\mathrm{m}$ ($0.01\,\mathrm{m}$ on the ankles and toes) and $\delta^{\mathrm{mov}}_{j}{=}0.05\,\mathrm{m}$ against moving obstacles ($0.03\,\mathrm{m}$ on the legs and toes); the safety radius $\rho{=}0.10\,\mathrm{m}$; and Huber $\delta_H{=}1$. The latent is initialized at $\sigma_{\mathrm{init}}{=}0.1$ (Alg.~\ref{alg:noiseopt}), which we observe to produce more natural-looking results when using noise optimization; the baselines use $\sigma{=}1$. The edge term is enabled only where the scene carries a heightfield, and the hand-contact term only where the task carries a grip.

\begin{table}[t]
\centering
\footnotesize
\begin{tabular}{lcccccc}
\toprule
Task & $K$ & $\eta$ & $w_c$ & $w_f$ & $w_e$ & $w_h$ \\
\midrule
Chair (sit, stand)   &  50 & 0.01 & 0.1 & 0.1 & 0   & 0 \\
Stairs (up, down)    &  50 & 0.05 & 1.0 & 1.5 & 1.0 & 0 \\
Step up/down         &  50 & 0.05 & 2.0 & 1.5 & 1.0 & 0 \\
Slalom               & 100 & 0.05 & 2.0 & 1.5 & 0   & 0 \\
Pick and place       & 100 & 0.01 & 0.5 & 0.1 & 0   & 1.0 \\
\addlinespace[3pt]
\multicolumn{7}{@{}l}{\textit{Agent-authored (Appendix~\ref{app:agenttasks})}} \\
Crawl under          & 100 & 0.05 & 2.0 & 1.5 & 0   & 0 \\
Duck under           & 100 & 0.05 & 2.0 & 1.5 & 0   & 0 \\
Squeeze through gap  & 100 & 0.05 & 2.0 & 1.5 & 0   & 0 \\
Push a box           & 100 & 0.05 & 2.0 & 1.5 & 0   & 1.0 \\
\bottomrule
\end{tabular}
\caption{Optimizer settings and loss weights per task family, for the eight
evaluated tasks and the four agent-authored behaviors. $K$ is the number
of outer Adam iterations of Alg.~\ref{alg:noiseopt} and $\eta$ its learning rate;
$w_g{=}1.0$ throughout. The two pick tasks share one recipe, as do
the two chair tasks and the two stair tasks. The agent-authored tasks were all
generated with a single recipe, differing only in whether the task carries a
grip.}
\label{tab:lossweights}
\end{table}

\section{Policy learning}
\label{app:policy}

\subsection{Simulation setup and dynamics randomization}
\label{app:simsetup}

\paragraph{Simulator and robot.} Teachers and students are trained in Isaac Sim through the \texttt{Holosoma} training stack~\cite{holosoma}. Physics is stepped at $200\,\mathrm{Hz}$ with a control decimation of four, giving the $50\,\mathrm{Hz}$ control rate. The robot is a Unitree G1 with $29$ actuated degrees of freedom over $32$ bodies, driven by joint-space PD control with per-joint gains and the manufacturer's torque limits; the policy output is a residual on a fixed default pose.

\paragraph{Episodes and termination.} Teachers are trained with $4096$ parallel environments and students with $384$. Episodes last $10$--$15\,\mathrm{s}$, drawing a reference motion sequence uniformly from the task's batch at each reset. An episode terminates when the root drifts more than $0.5\,\mathrm{m}$ from the reference, when the projected gravity of the root differs from the reference by more than $0.8$, when any of four bodies (the two ankles and the two wrists) exceeds $0.25\,\mathrm{m}$ of tracking error, or, on object tasks, when the object's pose relative to the robot deviates by more than $0.25\,\mathrm{m}$ or $0.8\,\mathrm{rad}$.

\paragraph{Dynamics randomization.} The following are resampled per environment: robot friction ($0.3$--$1.6$ static, $0.3$--$1.2$ dynamic) and restitution ($0.0$--$0.5$); object friction ($0.1$--$0.6$), restitution ($0.0$--$1.0$), mass ($1$--$4\times$) and principal inertia ($0.5$--$1.5\times$); base center-of-mass offset ($\pm0.025$\,m longitudinally, $\pm0.05$\,m laterally and vertically); and a $\pm0.01\,\mathrm{rad}$ joint-position bias. Random pushes are applied every $1$--$3\,\mathrm{s}$ at up to $0.5\,\mathrm{m/s}$ linear and $0.78\,\mathrm{rad/s}$ angular velocity. Observations carry Gaussian noise on base angular velocity ($0.2$), joint position ($0.01$), joint velocity ($0.5$), and reference orientation ($0.05$). The teacher and student stages randomize identically; the student adds the depth randomization (Appendix~\ref{app:depthaug}).

\subsection{Motion-tracking teacher}
\label{app:teacher}

A single teacher (Sec.~\ref{sec:method:teacher}) is trained per task with PPO~\cite{schulman2017ppo} on the full batch of motions generated for that task: at each episode reset, a reference trajectory is sampled uniformly from the batch and the policy is required to track it. The actor consumes the robot's proprioception, the kinematic reference over a short horizon of future steps, and a privileged terrain height scan; on object tasks it additionally receives the object's pose, velocity, and sampled surface points in the robot frame. It outputs a $29$-D joint-angle residual over the default pose, applied through PD control. The critic is further privileged with reference-body and tracked-body poses and the base linear velocity. Rewards combine standard motion-tracking terms (global and per-body pose, position, and velocity tracking, plus object-pose tracking on object tasks) with regularizers on action rate, DoF limits, and undesired contacts. The actor-critic architecture and PPO hyperparameters are the default values from~\cite{holosoma}.

\subsection{Depth-conditioned flow student}
\label{app:student}

\paragraph{Objective.} Along the straight-line path $\mathbf{a}_t = (1-t)\,\mathbf{a}^{\mathrm{tea}} + t\,\boldsymbol{\epsilon}$ that connects the teacher action $\mathbf{a}^{\mathrm{tea}}$ at $t{=}0$ to Gaussian noise $\boldsymbol{\epsilon}\sim\mathcal{N}(\mathbf{0},\mathbf{I})$ at $t{=}1$, the student (Sec.~\ref{sec:method:student}) regresses a conditional velocity field $v_{\theta}$ onto the velocity of that path, given the onboard sensor context $\mathbf{c}$:
\begin{equation}
\label{eq:student}
\mathcal{L}_{\mathrm{student}}
= \mathbb{E}_{t,\boldsymbol{\epsilon}}\left\|\,
v_{\theta}(\mathbf{a}_t, t, \mathbf{c}) - \left(\boldsymbol{\epsilon} - \mathbf{a}^{\mathrm{tea}}\right)
\right\|^{2}.
\end{equation}
At test time an action is obtained by integrating $v_{\theta}$ from $t{=}1$ to $t{=}0$.

\paragraph{Observations.} The student (Sec.~\ref{sec:method:student}) observes only quantities the robot carries: a short history of head-mounted depth frames, each encoded by a shared depth CNN into a compact embedding, and a proprioception stream (base angular velocity, projected gravity, joint-angle deviations from the default pose, joint velocities, end-effector positions in the base frame, and the previous action) kept as both a short recent window and a longer buffer compressed by a temporal 1-D CNN. Shorter-horizon tasks use a smaller memory. It outputs the same $29$-D joint-angle residual as the teacher.

\paragraph{Architecture.} The velocity field $v_{\theta}$ of Eq.~\ref{eq:student} is a small transformer that attends over tokens for the proprioception history, the depth frames, the current action $\mathbf{a}_t$, and the flow time $t$ (alongside learnable register tokens), and projects the action token's output back to the $29$-D velocity.

\paragraph{Training.} Distillation is on-policy behavior cloning: the student rolls out in simulation and the frozen teacher relabels every visited state with its action, the regression target of Eq.~\ref{eq:student}; the teacher itself never acts in the environment ($\beta{=}0$ throughout). At evaluation and deployment the field is integrated with $5$ fixed Euler steps. A value critic is trained alongside by standard value regression~\cite{schulman2017ppo} but contributes no gradient to the student. Training uses AdamW (learning rate $5{\times}10^{-4}$) for $25$k iterations across $384$ parallel environments; depth randomization is applied throughout.

\subsection{Depth randomization for sim-to-real}
\label{app:depthaug}

Rendered depth is clean, whereas the on-board camera returns holes, invalid regions, self-occlusion from the robot's body, and slight mount misalignment. To close this gap we add depth randomization during distillation, following~\cite{visualmimic}. It acts on the preprocessed frame stack of Appendix~\ref{app:student} (normalized so that $+0.5$ is far or missing and $-0.5$ near) per minibatch, in two stages.

\paragraph{Camera-rotation jitter.} Each sample draws one in-plane rotation $\alpha\sim\mathcal{U}(-5^\circ,5^\circ)$, applied to every frame of that sample's stack, modeling a fixed but uncertain mount orientation rather than per-frame shake.

\paragraph{Random masking.} Two families of occlusion are added. Fixed blocks anchored to the bottom corners of the frame model self-occlusion by the robot's own body; they are drawn per sample and held fixed across the stack, since the body stays in view over the whole window. Random rectangles model transient holes and dropouts; these are drawn independently per frame. Randomization is applied only during training and not during deployment.

\end{document}